\documentclass[acmsmall]{acmart}

\AtBeginDocument{%
  }

\usepackage{booktabs}
\usepackage{mathrsfs}

\usepackage{graphicx}
\usepackage{amsmath}

\usepackage{booktabs}
\usepackage{subfigure}
\usepackage{caption}
\usepackage{subcaption}
\usepackage{multirow}
\usepackage{colortbl}

\setcopyright{acmcopyright}
\copyrightyear{2023}
\acmYear{2023}
\acmDOI{XXXXXXX.XXXXXXX}

\acmJournal{JACM}
\acmVolume{37}
\acmNumber{4}
\acmArticle{111}
\acmMonth{8}

\begin{document}

\title{Ranking the Transferability of Adversarial Examples}

\author{Mosh Levy}
\email{moshe0110@gmail.com}
\authornote{Work was done while at Ben-Gurion University}
\affiliation{%
  \institution{Bar-Ilan university}
  \streetaddress{5290002}
  \city{Ramat-gan}
  \country{Israel}
}

\author{Guy Amit}
\email{guy5@post.bgu.ac.il}

\author{Yuval Elovici}
\email{elovici@bgu.ac.il}

\author{Yisroel Mirsky}
\email{yisroel@bgu.ac.il}
\authornote{Corresponding author}
\affiliation{%
  \institution{Ben-Gurion University}
  \streetaddress{1 Ben-Gurion Ave.}
  \city{Beersheba}
  \country{Israel}
}

\renewcommand{\shortauthors}{Levy et al.}

\begin{abstract}
Adversarial transferability in black-box scenarios presents a unique challenge: while attackers can employ surrogate models to craft adversarial examples, they lack assurance on whether these examples will successfully compromise the target model. Until now, the prevalent method to ascertain success has been trial and error—testing crafted samples directly on the victim model. This approach, however, risks detection with every attempt, forcing attackers to either perfect their first try or face exposure.

Our paper introduces a ranking strategy that refines the transfer attack process, enabling the attacker to estimate the likelihood of success without repeated trials on the victim's system. By leveraging a set of diverse surrogate models, our method can predict transferability of adversarial examples. This strategy can be used to either select the best sample to use in an attack or the best perturbation to apply to a specific sample.

Using our strategy, we were able to raise the transferability of adversarial examples from a mere 20\%—akin to random selection—up to near upper-bound levels, with some scenarios even witnessing a 100\% success rate. This substantial improvement not only sheds light on the shared susceptibilities across diverse architectures but also demonstrates that attackers can forego the detectable trial-and-error tactics raising increasing the threat of surrogate-based attacks.
\end{abstract}




\maketitle

\section{Introduction}
\label{sec:intro}
Neural networks are vulnerable to adversarial examples in which adversaries aim to change the prediction of a model $f$ on an input $x$ in a covert manner \cite{szegedy2014intriguing}. The common form of this attack is where an adversarial example $x'=x+\delta$ is created such that $f(x+\delta) \neq f(x)$ where $||\delta||<\epsilon$. In other words, the adversarial example changes the model's prediction yet the $x'$ appears the same as $x$. 

Many popular and powerful adversarial attacks (such as PGD \cite{madry2018towards} and CW \cite{carlini2017towards}) are whitebox attacks. This means that in order to use these algorithms to generate $x'$, the attacker must have access to the learnt model parameters in $f$ (i.e., the neural network's weights). Although this may seem like a strong limitation for attackers, it has been shown that different neural networks can share the same vulnerabilities to an adversarial example \citep{szegedy2014intriguing,tramer2017space}. As such, an attacker can simply train a surrogate model $f'$ on a similar dataset \cite{zhou2020dast}, attack $f'$ to generate $x'$, and then deploy $x'$ on the blackbox model $f$ knowing that there will be a decent probability of success. This attack is called a transfer attack  \citep{papernot2016transferability}. This type of attack has been found to be effective across models trained on different subsets of the data \citep{szegedy2014intriguing}, across domains \citep{naseer2019cross} and even across tasks \citep{xie2017adversarial}.  

\begin{figure}[t]
    \centering
    \includegraphics[width=0.8\textwidth]{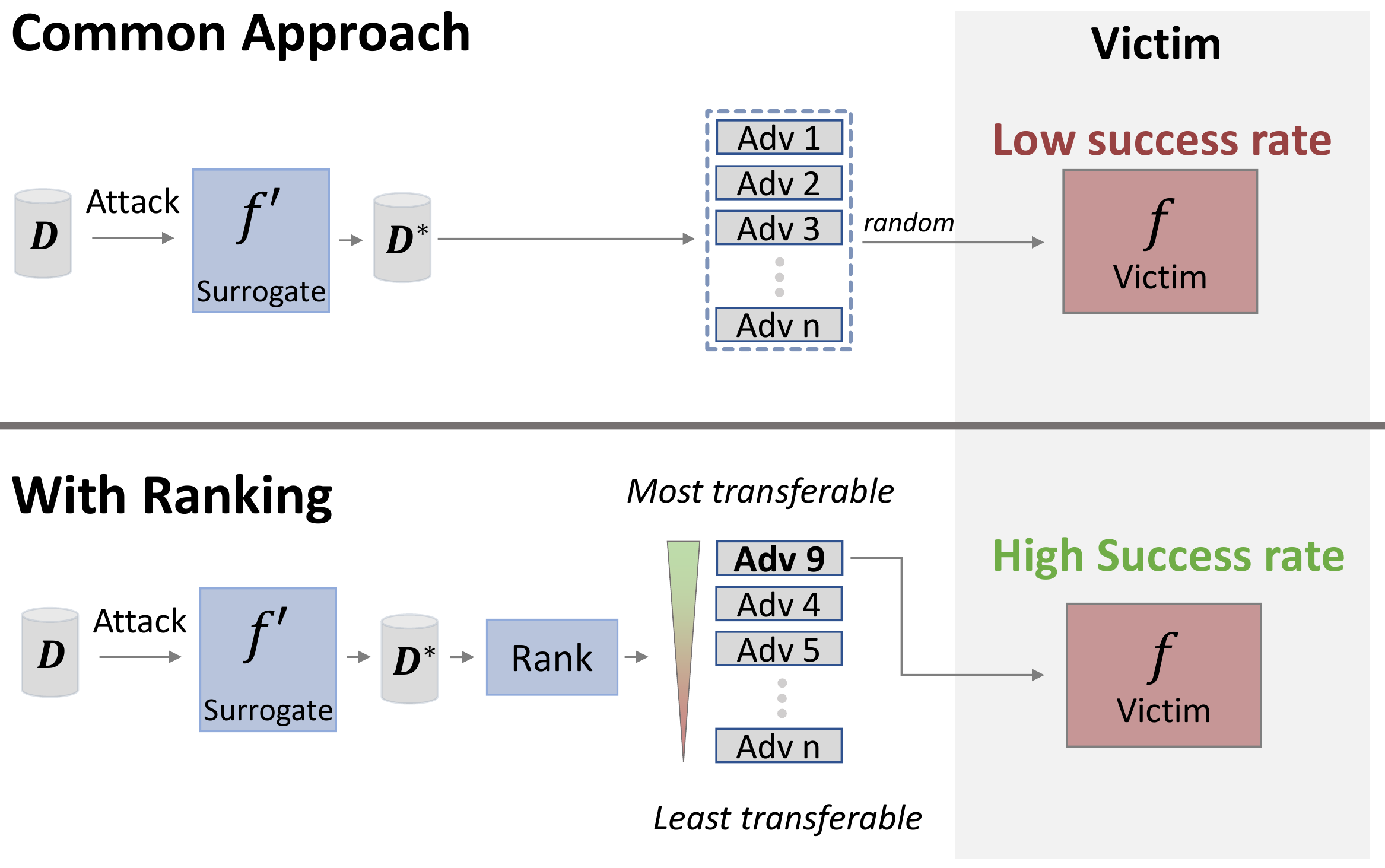}
    \caption{An illustration showing how ranking is beneficial to an adversary who is targeting model $f$ in a blackbox setting. Here $\mathscr{D}$ is the dataset available to the attacker and $\mathscr{D}^*$ are adversarial examples. $f$ is not available to the adversary.}
\label{firstpage}
\end{figure}

In the field of adversarial machine learning, understanding transferability is essential, especially when an attacker has only one or a few chances to succeed. In real-world situations, like trying to trick a facial recognition system at an airport or fooling a fraud detection algorithm at a bank, the attacker doesn't have the luxury to keep trying until they succeed. If they fail, the system might lock them out or increase security, making it even harder to try again. Therefore, it's important for an attacker to choose the adversarial example that is most likely to work on the first try. This is why ranking adversarial examples based on how well they are expected to transfer and fool the target model is crucial—it increases the odds of success in situations where there's no room for error (see Fig. \ref{firstpage}).

In some attacks, the adversary may search for the best sample(s) to use and use ranking to select them ($x_i\in D$ for $x'_i=x_i+\delta$). However, in other attacks, the adversary is limited to use one or more specific samples (such as an image of a individual). In these attacks an adversary could generate several adversarial examples for each given sample and then use ranking to select the best \textit{perturbation} ($\delta_j$ for $x'_j=x_j+\delta_j$). 

To understand these cases, let's consider two example scenarios: In the first scenario, an attacker may be trying to evade detection of some anti-virus model $f$ and can select one malware from a set for this purpose (represented as $x_i$) such that some modification to it ($\delta$) will make it perceived by $f$ as benign software \cite{mahdavifar2019application,grosse2017adversarial}. 
Here the ranking is done on $x\in \mathscr{D}$ since the attacker can convert any of the malwares into an adversarial example, so the attacker will choose the malware that is best for transferability. 
In the second scenario, an attacker may want to tamper a specific patient's medical image $x$ with some perturbation $\delta_j$ such that $x'_j$ will be falsely classified as containing some medical condition \cite{hirano2021universal,levy2022security}. 
Here ranking is done on potential perturbations because there is only one $x$. We note that in both cases, the attacker (1) has only one attempt to avoid being caught or (2) cannot get feedback from $f$, but must select the sample $x_i$ or perturbation $\delta_j$ which will most likely transfer to the victim's model $f$.

To find the best $x_i$ or $\delta_j$ using surrogate $f'$, the attacker must rank potential adversarial examples accordingly to their expected attack success on $f$. We call this measure the \textit{expected transferability} (ET). To the best of our knowledge, there are no works which propose a means for ranking adversarial examples according to their ET. Current works, such as \cite{tramer2017space,dong2019evading,zhu2021rethinking}, determine if $x'$ transfers by directly evaluating it on the victim's model $f$. However, in a blackbox setting, an attacker cannot use $f$ to measure success. Therefore, this approach can only be used as an upper bound, but cannot be used to (1) help the attacker select the best adversarial example(s) or (2) measure a model's robustness to transferability attacks given the attacker's limitations. 



In this paper, we explore the topic of ranking adversarial examples according to their ET. Our work offers several contributions: (1) we propose the concept of expected transferability and define the ranking problem for adversarial examples, (2) we suggest a way to approximate the ET of an adversarial example and a heuristical way to increase the accuracy and practicality of the method, (3) we introduce a new metric (``transferability at $k$'') to measure attack performance considering an attacker's best efforts and (4) we frame the problem of transferability realistically in the perspective of a blackbox attacker: we propose the use of additional surrogates to evaluate transferability.

\section{Definitions}
In this section we introduce the concept of expected transferability, define the task of ranking an adversarial example's transferability, and propose the metric ``transferability at $k$''.

\subsection{Expected Transferability (ET)}
In the domain of transferability, where the attacker is positioned in uncertainty, a deterministic answer as to whether an adversarial example will succeed (fool the victim) is impossible. The attacker is faced against an unknown victim model, which by definition the attacker has incomplete information about. An appropriate way to consider the victim model is as if it was sampled from the pool of all possible victim models. Therefore, rather than a guarantee, the attacker is interested in the expectation of each adversarial example to be successful. 

To define formally the Expected Transferability the attacker is interested in we will need to define what is the pool of possible victim models.
Let $F$ be the set of all possible models scoped and based on the attacker's knowledge of $f$ (i.e., $F$ is the set of all surrogate models that reflect $f$). 
In our setting, the attacker uses the surrogate model $f'\in F$ to create adversarial examples (denoted as the set $\mathscr{D}^*$).

We define the expected transferability of an adversarial example $x'_i\in \mathscr{D}^*$ as the probability that $x'_i$ will successfully transfer to a random model in $F$. A successful transfer of $x'_i$ to model $f_j\in F$ can be defined as the case where $f_j(x'_i)\neq y_i$ for an untargeted attack where $y_i$ is the ground truth label of $x_i$.\footnote{
In this work, we concentrate on untargeted attacks, though our methodology could be adapted for targeted attacks with the aim of achieving $f_j(x'_i)= y_t$ where $y_t$ represents the class that the attacker intends to mimic.} Following the notation convention used for adversarial examples (e.g., \cite{tramer2020adaptive}), the symbol $\neq$ should be interpreted in the context of classification outcomes rather than as a strict boolean operation. Specifically, it indicates that the probability distribution output by $f(x')$ does not assign the highest probability to the ground truth class $y$, meaning that the class with the highest probability in $f(x')$ is different from $y$.

It can be said that the attacker's goal is to select a sample $x'_i$ which has the highest probability to transfer to a random model drawn from the population $F$. For untargeted attacks, we can measure $x'_i$'s transferability with 
\begin{equation} \label{eq:transferability}
    S(x'_i)=\mathop{\mathbb{E}}_{f\sim F}[f(x'_i)\neq y_i]
\end{equation}
$S$ can be used to rank adversarial examples because if $S(x'_i)>S(x'_j)$, then $x'_i$ is more likely to transfer to a random model in $F$ than $x'_j$. Note that (\ref{eq:transferability}) can similarly be defined for targeted attacks as well.

\subsection{Transferability Ranking}
Given $S$, the attacker can sort the potential adversarial examples according to their ET.
Therefore, we define the task of transferability ranking as the problem of obtaining an ordered set of adversarial examples $\{x'_1,x'_2,...\}$ such that $x'_i \in \mathscr{D}^*$ and $x'_i$ is more likely to transfer than $x'_j$ if $S(x'_i)>S(x'_j)$. 

Note that when applying $S(x'_i)$, it is possible to measure the expected transferability rank of different samples from a dataset $x'_i=x_i +\delta, x_i \in \mathscr{D}$ or different perturbations on a specific same sample from the dataset $x'_i=x_i+\delta_j, x \in \mathscr{D}, \delta_j\in \delta$.
Ranking adversarial examples by perturbation is relevant for attacks where multiple runs of the attack algorithm produce different perturbations \cite{madry2018towards}.

\subsection{Transferability at $k$}
In a real-world attack, an attacker will curate a finite set of $k$ adversarial examples on $f'$ to use against $f$. To ensure success, it is critical that the attacker select the $k$ samples that have the highest ET scores. 

The top $k$ samples of $\mathscr{D^*}$ are denoted as the set  $S_k(\mathscr{D^*})$ where setting $k=1$ is equivalent to selecting the sample that is the most likely to transfer. 

Identifying the top $k$ samples is not only useful for the attacker, but also the defender. This is because a defender can evaluate his or her model's robustness to attacks given the attacker's best efforts (attacks using the top $k$ samples). We call this performance measure the \textit{transferability at $k$} defined as
\begin{equation}\label{eq:tatk}
    T_k(\mathscr{D^*})=\frac{1}{k}\sum_{x'\in S_k(\mathscr{D^*})} \left( f(x')\neq y \right)
\end{equation}
which is the average number of cases where the top $k$ samples selected by $S$ successfully transferred to the victim $f$.

This evaluation expresses a wide variety of use cases, some cases call for the use of only a small amount of adversarial samples, while other cases require large amounts so the attacker will only be interested in the small amount of worst adversarial attacks for transferability so they can be omitted. Note that the score of a specific $k$ is bounded by success of the $k$ most transferable samples in the dataset.
As such, when $K=|\mathscr{D^*}|$ the score will be the average attack success rate of the adversarial samples in the dataset for any ranking of the samples.


\section{Implementation} \label{method}
In this section we propose methods for implementing $S$ and estimating the \textit{transferability at $k$} without access to $f$.

\subsection{Approximate Expected Transferability (AET)}
Although the set $F$ is potentially infinite, we can approximate it by sampling models from the population $F_0 \subset F$. With $F_0$ we can approximate $S$ by computing
\begin{equation}
\label{baseline_eq}
    S(x'_i)=\frac{1}{|F_0|}
    \sum_{j=1}^{|F_0|} \left(f_j(x'_i)\neq y_i\right) 
\end{equation}
for $f_j\in F_0$.

In summary, we propose the use of multiple surrogate models to estimate ET: one surrogate model is used to generate the adversarial example ($f'\in F$) and one or more surrogate models ($F_0 \subset F$) are used to estimate the transferability of the adversarial example to $f$.

\subsection{Heuristical Expected Transferability (HET)}
Although we can use (\ref{baseline_eq}) to compute ET, the approach raises a technical challenge: it is impractical to train a significantly large set of surrogate models $F_0$. For example, training a single Resnet-50 on ImageNet can take up to 4 days using common hardware\cite{wightman2021resnet}. However, if $|F_0|$ is too small then $S$ will suffer from a lack of granularity. This is because, according to (\ref{baseline_eq}), each model reports a 0 or 1 if the attack fails or succeeds. To exemplify the issue of granularity, consider a case where $|F_0|=10$ and we set $k=100$. If $\mathscr{D}^*$ contains 1000 adversarial examples which fool all 10 models, then all $1000$ samples will receive a score of $1.0$. However, the true $S$ of these samples vary with respect to $F$. As a result, we will be selecting $k=100$ random samples randomly from these $1000$ which is not ideal. 

To mitigate this issue, we propose using continuous values to capture attack success for $x'$ on each model. Specifically, for each model, we use the model's confidence for the input sample's ground-truth class. This value implicitly captures how successful $x'$ is at changing the model's prediction since lower values indicate a higher likelihood that $x'$ will not be classified correctly \cite{GoodfellowSS14,madry2018towards}. When averaged across $|F_0|$ models, we can obtain a smoother probability which generalizes better to the population $F$. Averaging model confidences is a popular ensemble technique used to join the prediction of multiple classifiers together \cite{ju2018relative}. However, here we use it to identify the degree in which a sample $x'$ exploits a set of models together.

To implement this heuristic approach, we modify (\ref{baseline_eq}) to
\begin{equation}\label{eq:hueristic}
    S(x'_i)=\frac{1}{|F_0|}
    \sum_{j=1}^{|F_0|} \left(1- \sigma_y \left(  f_j(x'_i) \right) \right),  f_j\in F_0
\end{equation}
where $\sigma_i (y)$ returns the Softmax value of the logit corresponding to the ground-truth label $y$. 

\begin{figure}[t]
    \centering
    \includegraphics[width=0.7\textwidth]{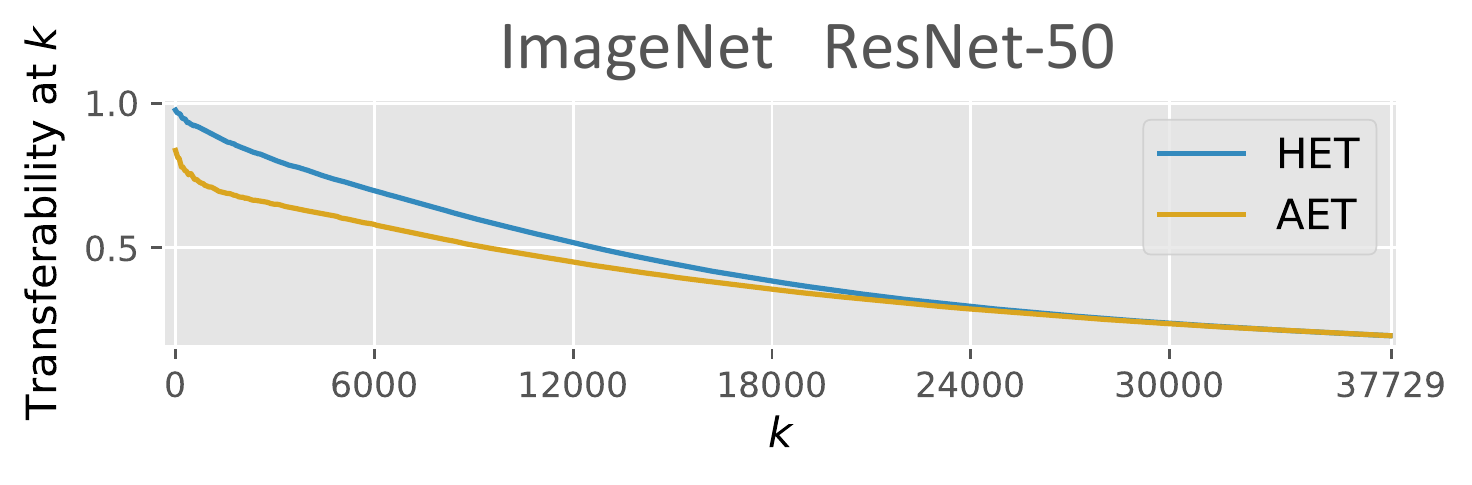}
    \caption{The ranking performance when using the heuristical expected transferability score (HET) compared to using the approximated expected transferability score (AET). Here both methods use $|F_0|=3$.}
    \label{baseline}
\end{figure}

We demonstrate the benefit of using HET (\ref{eq:hueristic}) over AET (\ref{baseline_eq}) with a simple experiment: We take a Resnet-50 architecture for both $f$ and $f'$, trained on the same ImageNet train set \cite{deng2009ImageNet}. Then, we create $\mathscr{D}^*$ by attacking the ImageNet test set with PGD ($\epsilon = \frac{1}{255}$). Finally, we compute the AET and HET on each sample in $\mathscr{D}^*$ with $|F_0|=3$ surrogates.\footnote{Note: $f$ is never included in $F_0$ for all of our experiments.} In Fig. \ref{baseline}, we plot the attack success rate of $\mathscr{D}^*$ on $f$ for different $k$ when sorting the samples according to AET and HET respectively. We observe that (1) although $\mathscr{D}^*$ has a 98\% attack success rate on $f'$, it only has a success rate of ~20\% on $f$ even though both $f$ and $f'$ are identical in design, and (2) HET performs better than AET, especially for lower $k$ (i.e., when we select the top ranked samples).





\subsection{Blackbox Ranking Strategies}

\begin{figure}[t]
    \centering
    \includegraphics[trim={0 4.8cm 0 0}, clip, width=0.7\textwidth]{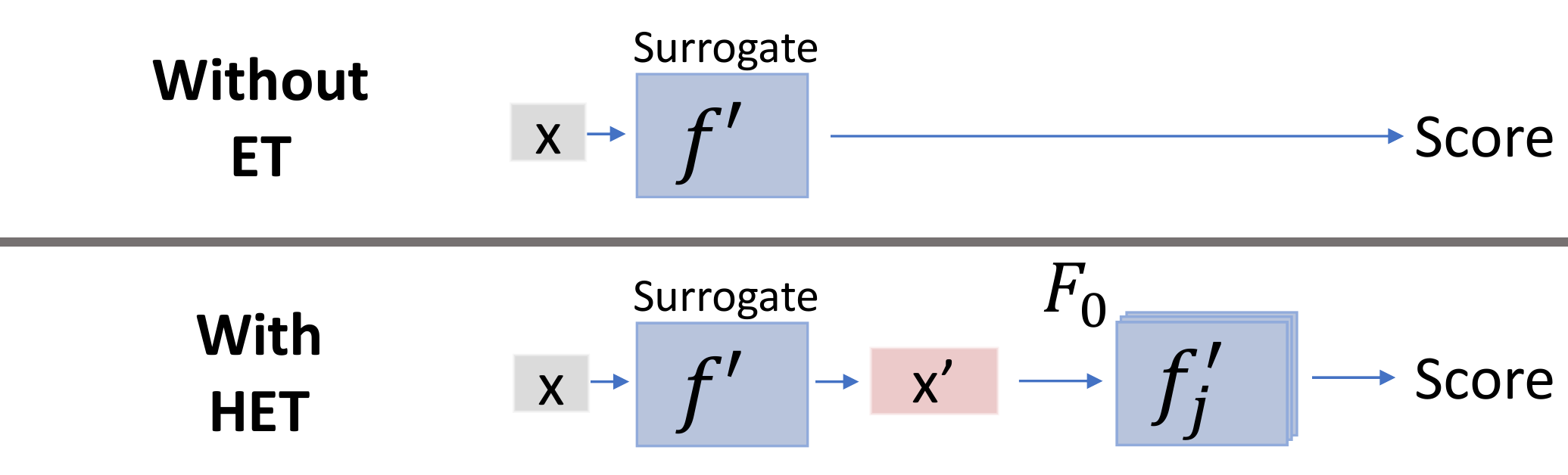}
    \caption{The proposed ranking strategy (HET) compared to the naive approach (without HET). Here \textit{x} is the input sample to be assigned a transferability score (the expected transferability). The auxiliary surrogates in $F_0$ reflect the victim's model $f$ depending on the attacker's knowledge of $f$ (architecture, training set, etc.)}
    \label{strategies}
\end{figure}

As discussed earlier, it is more likely that an attacker will measure a sample's transferability using surrogates and not the victim model $f$ (as done in previous works). Below, we propose two strategies for ranking the transferability of a sample $x'$ without using $f$ (illustrated in Fig. \ref{strategies}):

\begin{description}
    \item[Without ET.] This is the naive approach where the attacker uses one surrogate model ($f'$) to select the adversarial examples. There are a few ways of doing this. For example, the attacker can check if $x'$ successfully fools $f'$ and then assume that it will also work on $f$ because $f \in F$. Another way is to evaluate the confidence of $f'$ ($\sigma_i$) on the clean sample $x$ to identify an $x$ which will be easy to attack \cite{sourcesamples,zhu2021rethinking}. Although this strategy is efficient, it does not generalize well to $F$. Even in a blackbox setting, where the attacker knows the victim's architecture and training set, a sample $x'$ made on $f'$ will not necessarily work on $f$. It was shown that even when the only difference is the model's random initialization, predicting a specific sample's transferability is still a challenging problem \cite{katzir2021s}.  
    
    \item[With HET.] In this strategy, the attacker utilizes multiple surrogate models ($F_0$) to approximate the expected transferability of $x'$ to $f$, as expressed in (\ref{eq:hueristic}). Here, the performance depends more on the attacker's knowledge of $f$ (the variability of $F$) but less so on the random artifacts caused by initialization of weights and the training data used.\footnote{This assumes that the training data for each model in $F$ is drawn from the same distribution.}
    This is because the averaging mitigates cases where there are only a few outlier models in $F_0$ which are vulnerable to $x'$. As a result, the final transferability score captures how well $x'$ transfers to vulnerabilities which are common among the models in $F_0$. The concept of models having shared vulnerabilities has been shown in works such as \cite{wang2020generating,nakkiran2019a}. 
    
\end{description}

\section{Experiment Setup}
\label{setup}
In this section we present the experiments which we have performed to evaluate the proposed blackbox transferability ranking strategy. 

\subsection{Evaluation Measures} To evaluate our ranking methods, we use \textit{transferability at $k$} as defined in (\ref{eq:tatk}). Note that \textit{transferability at $k$} can also be viewed as the attack success rate on $f$ for the top-$k$ recommended samples. We remind the reader that ranking is performed without access to $f$ or knowledge of $f$. Therefore, it can be said that the \textit{transferability at $k$} measures the adversary's attack success rate in a black box setting when only $k$ attempts (attacks) are allowed. 

\subsection{Datasets} 
For our experiments, we used four datasets:
\begin{description}
    \item[CIFAR10.] An image classification dataset which contains 60K images from 10 categories having a resolution of 32x32.
    \item[ImageNet.] A popular image classification benchmark dataset containing about 1.2M images from 1000 classes downsamples to a resolution of 224x224.
    \item[RSNA-X-ray.] A medical dataset, containing approximately 30K pneumonia chest  X-ray images resized to a resolution of 224x224. The dataset contains three classes and was originally published on Kaggle\footnote{https://www.kaggle.com/competitions/rsna-pneumonia-detection-challenge/overview} by the Radiological Society of North America.
    \item[Road Sign.] A dataset for traffic sign classification containing 3K images from 58 different up-sampled to a  resolution of 224x224. The dataset was originally published on Kaggle\footnote{https://www.kaggle.com/datasets/ahemateja19bec1025/traffic-sign-dataset-classification?resource=download}. 

\end{description}

For the ImageNet and CIFAR10 datasets we have used the original data splits, whereas for the RSNA-X-Ray and Road\_Sign dataset, the images were split train, test, and validation sets with respective sizes of into 70\%, 20\% and 10\%.
The training sets were used to train $f$ and $f'$ and the test sets were used to create the adversarial examples ($\mathscr{D}^*$). 
Since $f(x)\neq y$ is counted as a successful attack, we must remove all samples from the test set where the clean sample is misclassified. This is done in order to avoid bias and focus our results strictly on samples which transfer as a result of the attack. 

\subsection{Architectures} 

In our experiments we used five different architectures:
\begin{description}

    \item[\texttt{DenseNet-121}.] Employs a dense connectivity pattern which connects each layer to every other layer in a feed-forward fashion. Its 121 layers are divided into dense blocks that ensure maximum information flow between layers, making it robust to perturbations.
    
    \item[\texttt{Efficientnet}.] Utilizes a compound scaling method that uniformly scales all dimensions of depth, width, and resolution with a fixed set of scaling coefficients. This architecture provides a balance between speed and accuracy, optimized to perform well even with limited computational resources.
    In CIFAR10 evaluations we use Efficientnet-b0 and for the reset we use Efficientnet-b2.
    
    \item[\texttt{Resnet18}.] Introduces skip connections to allow the flow of gradients through the network without attenuation. With 18 layers, it is relatively shallow, ensuring quick computations while still capturing complex features.
    
    \item[\texttt{Vision Transformer (ViT)}.] The Vision Transformer applies the principles of transformer models, primarily used in NLP, to image classification tasks. It treats image patches as sequences, allowing for global receptive fields from the outset of the model.
    For evaluations conducted on CIFAR10 we use a ViT model with 7 layers, 12 heads and an MLP dimension of 1152.
    For evaluations performed on other datasets, we used the ViT\_b\_16 architecture.
    
    \item[\texttt{Swin Transformer} (Swin\_s).] Implements a hierarchical transformer whose representations are computed with shifted windows, enabling efficient modeling of image data with varying scales and sizes.

\end{description}

In experiments conducted on X-Ray and Road\_Sign datasets, we fine-tuned pre-trained ImageNet models obtained from torchvision~\footnote{https://pytorch.org/vision/stable/models.html}. Conversely, for CIFAR10, we trained models from scratch. 
In the context of the ImageNet experiments, we employed the same pre-trained models as in the X-Ray and Road\_Sign experiments, with the sole exception being experiments necessitating the evaluation of identical model architectures.
In such cases, we utilized a larger variation of the same model also obtained from torchvision.

\subsection{Threat Model}
For our all of our experiments, we consider a black-box adversary that has no knowledge of the victim's architecture. To simulate this setting, we ensured that the architectures in used for $f$, $f'$ and those in $F_0$ were all unique. This simulates a black box setting because the architectures used by the adversary (surrogates $f'$ and $F_0$) will be different from architecture used in the victim model $f$.\footnote{The only exception is where the victim uses the ViT architecture. In this case alone we let the adversary uses a different sized ViT in $F_0$ due to time limitations in training.}

\subsection{Attack Algorithms} 
For the attacks, we use FGSM~\cite{goodfellow2014explaining}, PGD~\cite{madry2017towards} and PGD+Momentum (denoted as Momentum)~\cite{dong2018boosting}, which should have increased transferability according to~\cite{xie2019improving}. All of these algorithms are considered accepted baselines when evaluating adversarial attacks \cite{GoodfellowSS14,croce2020reliable,tramer2020adaptive}. The FGSM attack performs a single optimization step on $x$ to generate $\delta$. 
The PGD and Momentum algorithms perform multiple iterations where each iteration normalizes $\delta$ according to a given p-norm. 

In our experiments, we only perform untargeted attacks ($f(x') \neq y$), where the algorithm is executed on $f'$ alone (bounded by $\epsilon = \frac{1}{255}$ for CIFAR10 and $\epsilon = \frac{4}{255}$ ImageNet, X-Ray, and Road Sign datasets). In our experiments, we used only PGD unless explicitly stated otherwise.

\subsection{Ranking Algorithms}
We evaluate our three ranking strategies: 
\begin{description}
    \item[SoftMax] In this implementation of the strategy we score a sample's transferability by taking $1-\sigma_i(f'(x))$ where $x$ is the clean sample. The use of Softmax here is inspired from the works of \cite{sourcesamples} where Softmax is used to capture a model's instability in $f$ (not $f'$).
    \item[SoftMax+Noise] For this version we follow the work of \cite{zhu2021rethinking}. In their work the authors found that samples which are sensitive to noise on the victim model $f$ happen to transfer better from $f'$ to $f$. We extend their work to the task of ranking: each clean sample in the test set is scored according to how much random noise impacts the confidence of the surrogate $f'$. Samples which are more sensitive are ranked higher. Similar to \cite{zhu2021rethinking}, we also use Gaussian noise and set std=$\frac{16}{255}$.
    \item[HET (ours)] To implement HET, we use one surrogate model $f'$ to produce the adversarial examples and a set of three other unique surrogate models as $F_0$ to rank them.

\end{description}

Ranking with these strategies is achieved by (1) computing the respective score on each adversarial example $x' \in \mathscr{D}^*$ and then (2) sorting the samples by their score (descending order). 

As a baseline evaluation, we evaluate the average transferability rate across all the dataset (no ranking). This baseline can also be viewed as a kind of \textbf{lower-bound} on performance.
Note that this is essentially the same as the common transferability evaluation measure used in the literature. 
 
Finally, we contrast the above ranking methods to the performance of the optimal solution (\textbf{upper-bound}). In the task transferability ranking (blackbox), the optimal solution achieved by ordering the samples according to their performance on $f$ (as opposed to using surrogates).

\subsection{Environment \& Reproducibility}
Our code was written using Pytorch and all models were trained and executed on Nvidia 6000RTX GPUs. To reproduce our results, the reader can access our code online.\footnote{https://github.com/guyAmit/Adversarial\_Ranking}

\subsection{Experiments}

We investigate the following ranking tasks: (\textbf{Sample Ranking}) where the attacker must select the top $k$ samples from $\mathscr{D}$ to use in an attack on $f$, and (\textbf{Perturbation Ranking}) where the attacker must select the \textit{best} perturbation for a specific sample $x$ in an attack on $f$. In the sample ranking scenario we evaluate all value of $k$ from $k=1$ to $k=|\mathscr{D}^*|$. For the perturbation ranking scenario we set $k$ to be 5\%, 10\% and 20\% of the respective dataset size.

We performed the following experiments to evaluate how well our ranking strategies perform and generalize to different settings:

\begin{description}
    \item[E1 - Sample Ranking.] In these experiments we explore the performance of the ranking strategies in the context of ranking samples. In other words, given a set of $|\mathscr{D}|$ different images, if only $k$ can be used in an attack, which $k$ images should be selected to maximize likelihood of success (transferability).  
    
\begin{description}
    \item[E1.1 - Architectures.] The purpose of this experiment is to evaluate the generalization of the ranking strategies to different blackbox settings. In this experiment, we explore the transferability of the ranked samples for every combination of surrogate and victim model architecture and generate adversarial examples on $f'$ using PGD. For each dataset and combination of architectures we evaluated the \textit{transferability at $k$} of the strategies for every possible value of $k$ (from $k=1$ to $k=|\mathscr{D}^*|$).

    \item[E1.2 - Attacks.] The objective of this experiment is to see if the ranking strategies generalize to different attacks and whether there are some attacks that transfer better than others. For each dataset and combination of architectures we evaluated the \textit{transferability at $k$} of the strategies where $k$ was set to 5\%, 10\% and 20\% of the respective dataset size.
\end{description}

\item[E2 - Pertubation Ranking.] In this experiment, we evaluate how well the strategies can be used to rank perturbations instead of samples. This experiment captures the setting where the adversary must use a specific sample (i.e., image) in the attack but can improve the likelihood of transferability by selecting the best perturbation for that sample.
\begin{description}
    \item[E2.1 - Performance.]  The experiment was conducted as follows: for each sample $x$, we generated 27 different perturbations using PGD with random starts and random values for alpha (between $\frac{0.1}{255} \text{ and } \frac{0.3}{255}$) and the number of iterations (between $10 \text{ and } 20$). We then ranked $x$ with each of these perturbations and took the highest ranked sample ($k=1$). This process was repeated 100 times per dataset. In a followup experiment, we then ranked these images setting $k$ to 5\%, 10\% and 20\% of the respective dataset size.
\end{description}
\end{description}

\label{results}

\section{Experiment Results}

\subsection{E1 - Sample Ranking}
\subsubsection{E1.1 - Architectures.}

\begin{figure}
     \centering
     \begin{tabular}{c}
          \includegraphics[width=0.92\textwidth,]{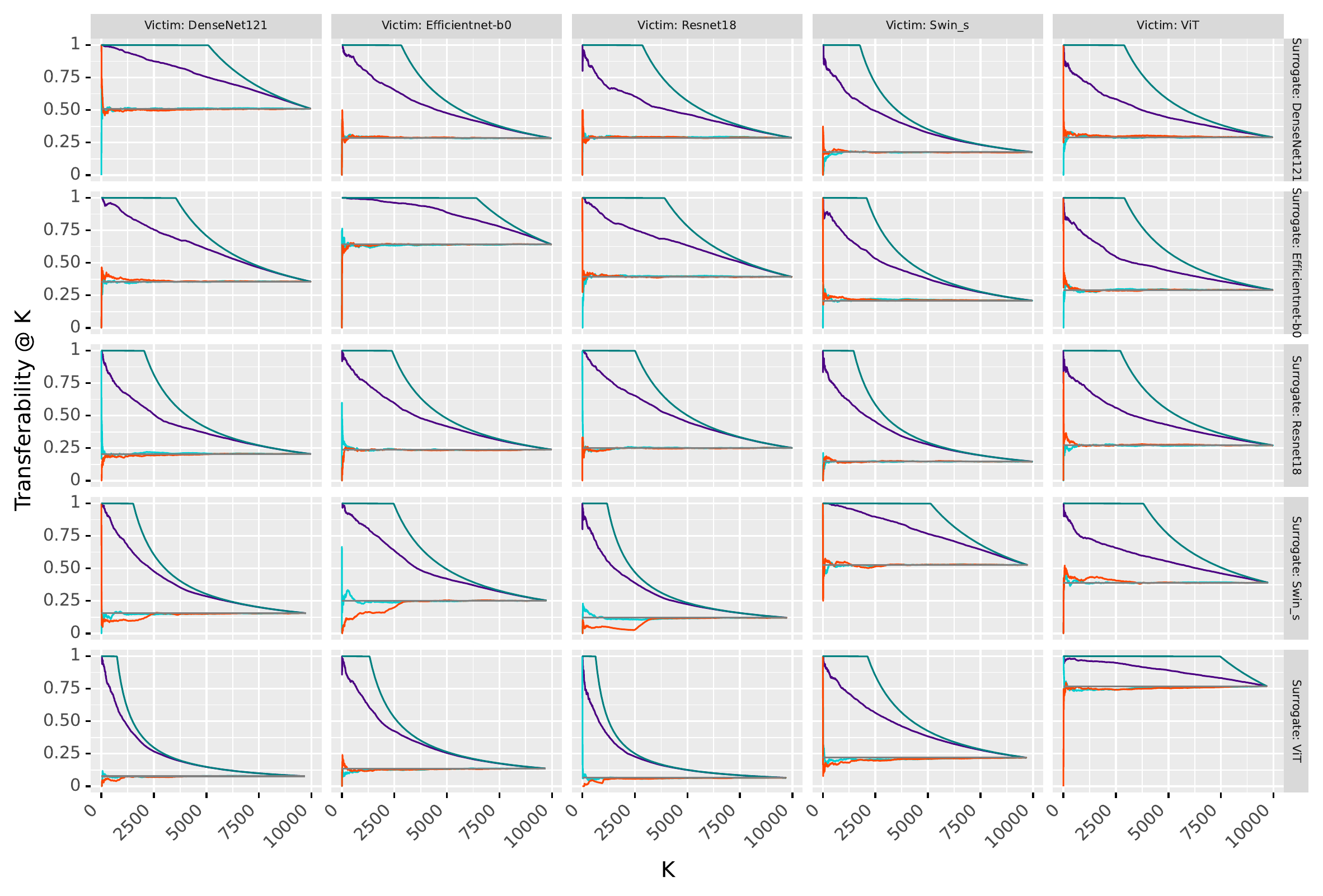} \\ \includegraphics[width=0.92\textwidth]{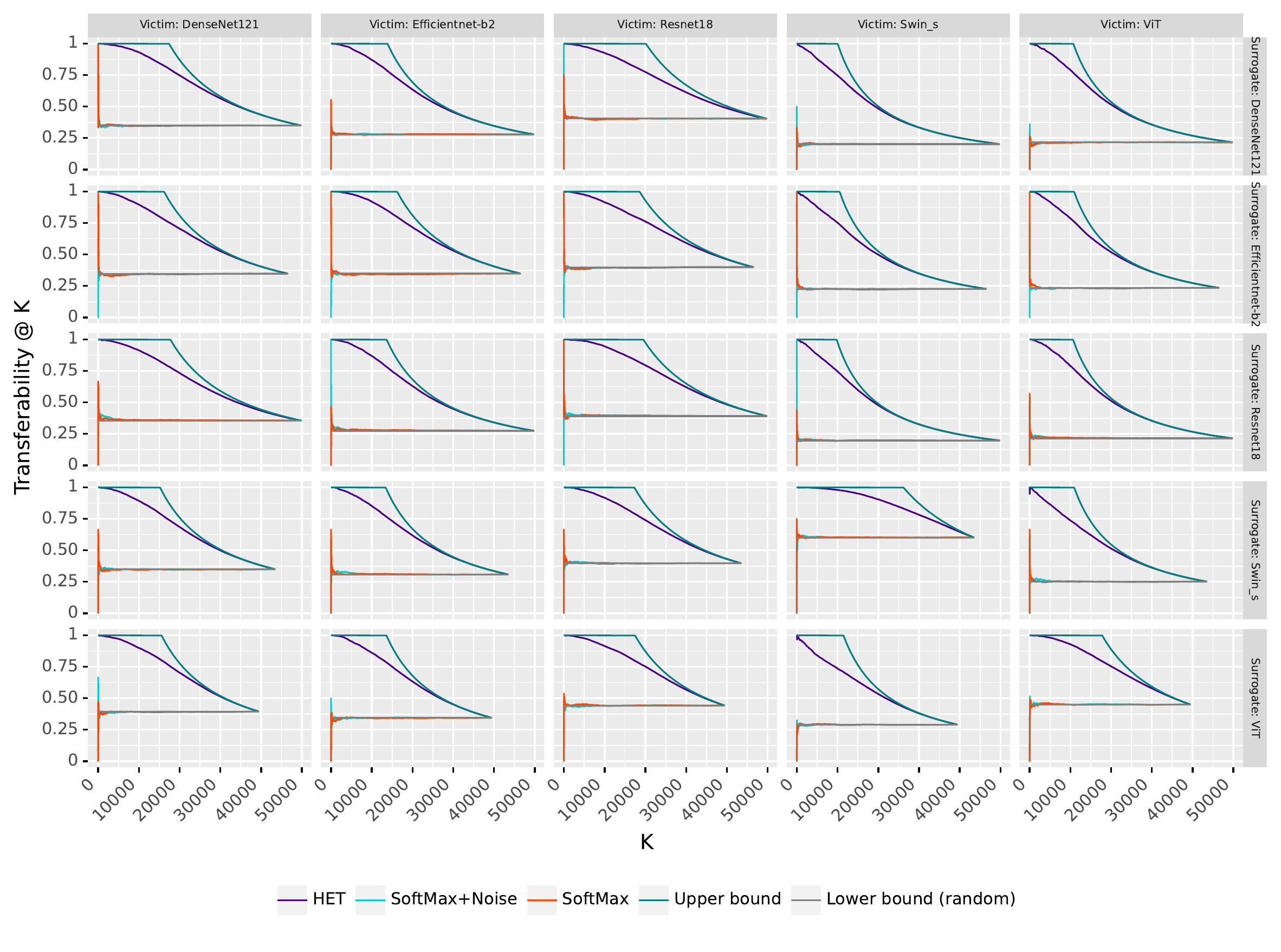} 
          
     \end{tabular}
   \caption{E1.1 Results - The performance of ranking strategies for the CIFAR10 (top) and ImageNet (bottom) datasets. Each cell plots the \textit{transferability at $k$} success rate for adversarial examples when ranked using different strategies across varied surrogate and victim model architectures. Columns represent the victim model architecture, and rows correspond to the surrogate model architecture.}
   \label{fig:E1.1A}
\end{figure}

\begin{figure}
     \centering
     \begin{tabular}{c}
          \includegraphics[width=0.92\textwidth,]{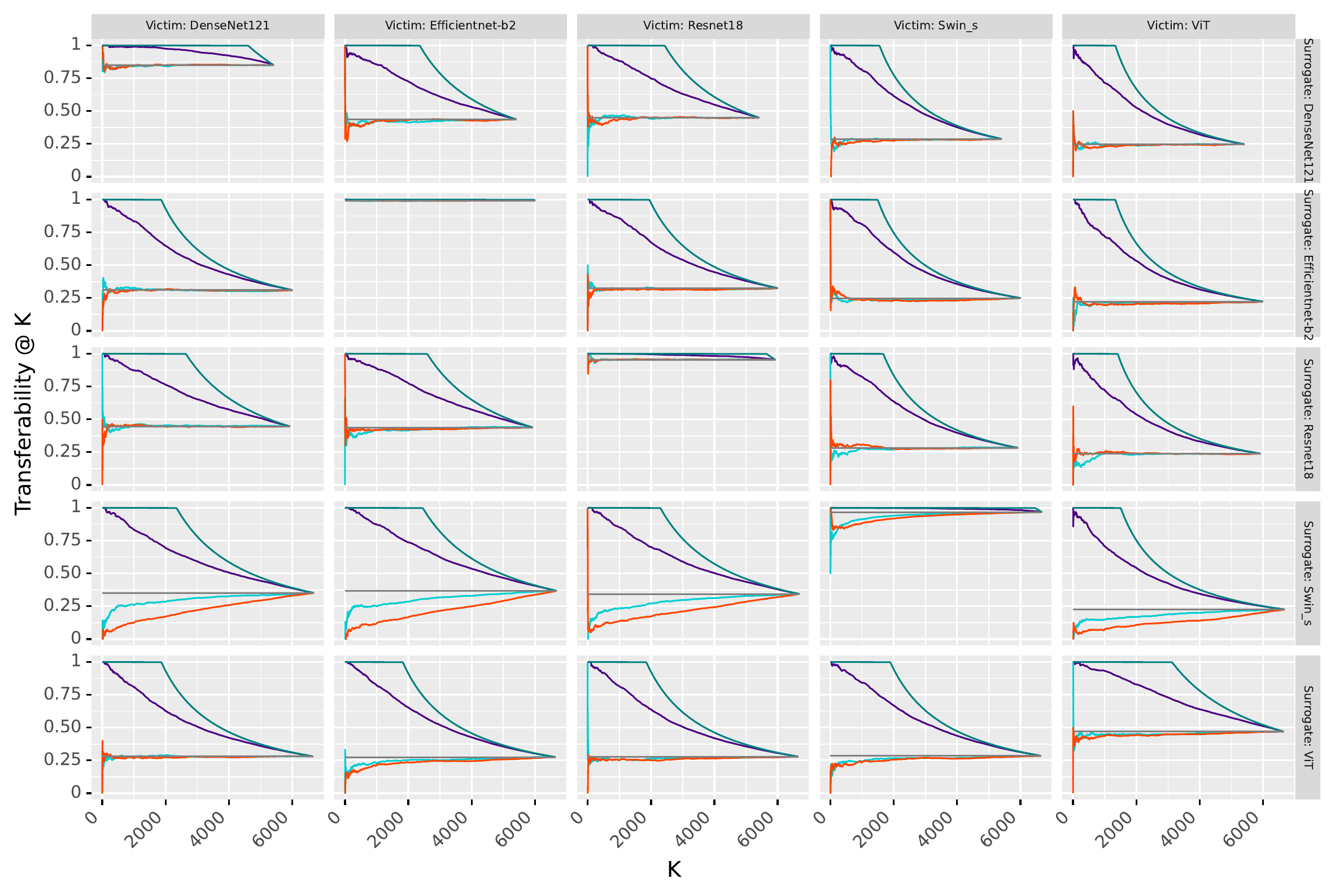} \\ \includegraphics[width=0.92\textwidth]{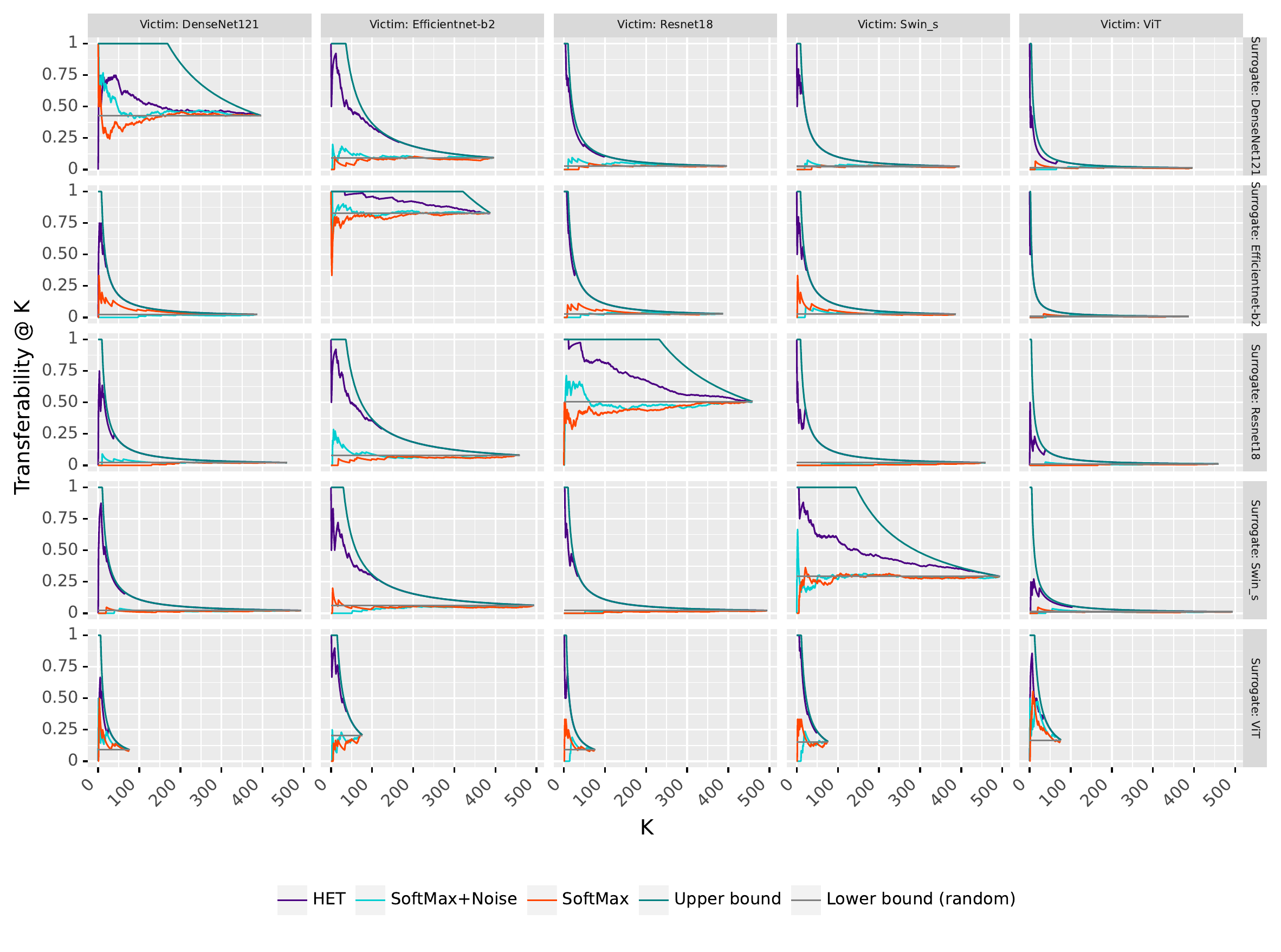} \\
     \end{tabular}
   \caption{E1.1 Results - The performance of ranking strategies for the X-Ray (top) and Road Sign (bottom) datasets. Each cell plots the \textit{transferability at $k$} success rate for adversarial examples when ranked using different strategies across varied surrogate and victim model architectures. Columns represent the victim model architecture, and rows correspond to the surrogate model architecture.}
   \label{fig:E1.1B}
\end{figure}

We direct the reader's attention to the results presented in Figures \ref{fig:E1.1A} and \ref{fig:E1.1B}, which presents the performance of our proposed ranking strategy, HET, across various datasets and model architecture pairings. Figure \ref{fig:E1.1A} details the outcomes for CIFAR10 and ImageNet, while Figure \ref{fig:E1.1B} delves into the X-Ray and Road Sign datasets. These figures plot the \textit{transferability at \(k\)} for a spectrum of \(k\) sizes, with the columns representing the victim architecture and the rows indicating the surrogate architecture utilized.

Our observations reveal a consistent trend across all datasets and architecture combinations: HET closely tracks the upperbound line, which represents the theoretical maximum transferability. Particularly for small values of \(k\), HET demonstrates a high likelihood of successful transferability, often achieving near-certain effectiveness. As \(k\) increases, HET maintains commendable performance, deviating by at most approximately 20

Conversely, the baseline methods, namely SoftMax and SoftMax+Noise, exhibit subpar performance. Out of 100 architecture combinations, only 45 yield a ranking that can be considered beneficial, and this is limited to instances where \(k\) is minimal (\(k=1\)). Moreover, the inconsistent performance between the two baseline methods presents an additional challenge, as it is unpredictable which method will be successful in any given scenario.

We note that the attacker only achieves the ideal results when he or she happens to select the same architecture for $f'$ as the victim in $f$ (captured by the diagonal of the figures).  Under these circumstances, transferability is notably higher for all methods across all \(k\) values, although not perfect. The discrepancy, even when architectures match, can be attributed to different training seeds affecting the models' decision boundaries, as discussed in \cite{katzir2021s}. Despite this, HET still significantly enhances transferability for nearly every combination of architectures (where the attacker guesses wrong). This is especially apparent in the setting where the attacker only needs to send one or just a few adversarial examples (low $k$) from the set of all potential images. 

When dissecting performance relative to the datasets, HET exhibits robust results for ImageNet, CIFAR10, and X-Ray. Nonetheless, there are instances within the Road Sign dataset where HET does not perform optimally at lower \(k\) values but recovers effectiveness at higher \(k\)s. This could be attributed to the varying image sizes in the dataset, which, when resized to fit the model input, may adversely impact the feature representations.

In summary, our comprehensive evaluation underscores the efficacy of the HET ranking strategy in diverse black-box settings, confirming its potential to improve adversarial example transferability in real-world attack scenarios.

\subsubsection{E1.2 - Attacks.}
\begin{table}[]
\resizebox{\linewidth}{!}{%

\begin{tabular}{|ccc|l|lll|lll|lll|lll|}
\hline
\multicolumn{3}{|c|}{}                                                                                                         & \multicolumn{1}{c|}{}                           & \multicolumn{3}{c|}{PGD}                                                                      & \multicolumn{3}{c|}{Momentum}                                                                 & \multicolumn{3}{c|}{FGSM}                                                                     & \multicolumn{3}{c|}{Upper B.}                                                                 \\ \cline{5-16} 
                                                  & \multicolumn{1}{c}{Surrogate}                          & Victim           & \multicolumn{1}{c|}{\multirow{-2}{*}{Lower B.}} & \multicolumn{1}{c}{5\%}       & \multicolumn{1}{c}{10\%}      & \multicolumn{1}{c|}{20\%}     & \multicolumn{1}{c}{5\%}       & \multicolumn{1}{c}{10\%}      & \multicolumn{1}{c|}{20\%}     & \multicolumn{1}{c}{5\%}       & \multicolumn{1}{c}{10\%}      & \multicolumn{1}{c|}{20\%}     & \multicolumn{1}{c}{5\%}       & \multicolumn{1}{c}{10\%}      & \multicolumn{1}{c|}{20\%}     \\ \hline
\multicolumn{1}{|c|}{}                            & \multicolumn{1}{c|}{}                                   & ViT\_l\_16       & \cellcolor[HTML]{FBA276}0.451                   & \cellcolor[HTML]{6AC07C}0.997 & \cellcolor[HTML]{85C87D}0.987 & \cellcolor[HTML]{D7E082}0.954 & \cellcolor[HTML]{6FC27C}0.995 & \cellcolor[HTML]{86C97E}0.986 & \cellcolor[HTML]{EBE683}0.946 & \cellcolor[HTML]{63BE7B}1.000 & \cellcolor[HTML]{6BC17C}0.997 & \cellcolor[HTML]{91CC7E}0.982 & \cellcolor[HTML]{63BE7B}1.000 & \cellcolor[HTML]{63BE7B}1.000 & \cellcolor[HTML]{63BE7B}1.000 \\
\hline
\end{tabular}
}
\caption{The comparative performance of HET across different attack algorithms and architecture combinations for the ImageNet and CIFAR10 datasets. Columns categorize the various attack algorithms employed, while rows detail the architecture pairings, with surrogate models ($F_0$) distinct from the victim's architecture.}
\label{tab:attacks_A}
\end{table}
\begin{table}[]
\resizebox{\linewidth}{!}{%

\begin{tabular}{|ccc|l|lll|lll|lll|lll|}
\hline
                                                   &                                                         &                  & \multicolumn{1}{c|}{}                           & \multicolumn{3}{c|}{PGD}                                                                      & \multicolumn{3}{c|}{Momentum}                                                                 & \multicolumn{3}{c|}{FGSM}                                                                     & \multicolumn{3}{c|}{Upper B.}                                                                 \\ \cline{5-16} 
                                                   & Surrogate                                               & Victim           & \multicolumn{1}{c|}{\multirow{-2}{*}{Lower B.}} & \multicolumn{1}{c}{5\%}      & \multicolumn{1}{c}{10\%}     & \multicolumn{1}{c|}{20\%}    & \multicolumn{1}{c}{5\%}      & \multicolumn{1}{c}{10\%}     & \multicolumn{1}{c|}{20\%}    & \multicolumn{1}{c}{5\%}      & \multicolumn{1}{c}{10\%}     & \multicolumn{1}{c|}{20\%}    & \multicolumn{1}{c}{5\%}      & \multicolumn{1}{c}{10\%}     & \multicolumn{1}{c|}{20\%}    \\ \hline
\multicolumn{1}{|c|}{}                             & \multicolumn{1}{c|}{}                                   & ViT      & \cellcolor[HTML]{FBB379}0.469                   & \cellcolor[HTML]{79C57D}0.976 & \cellcolor[HTML]{96CD7E}0.943 & \cellcolor[HTML]{BCD881}0.900 & \cellcolor[HTML]{7BC57D}0.973 & \cellcolor[HTML]{94CD7E}0.944 & \cellcolor[HTML]{BDD881}0.899 & \cellcolor[HTML]{7CC67D}0.972 & \cellcolor[HTML]{88C97E}0.958 & \cellcolor[HTML]{ACD380}0.918 & \cellcolor[HTML]{63BE7B}1.000 & \cellcolor[HTML]{63BE7B}1.000 & \cellcolor[HTML]{63BE7B}1.000 \\
\hline
\end{tabular}
}
\caption{The comparative performance of HET across different attack algorithms and architecture combinations for the X-Ray and Road Sign datasets. Columns categorize the various attack algorithms employed, while rows detail the architecture pairings, with surrogate models ($F_0$) distinct from the victim's architecture.}
\label{tab:attacks_B}
\end{table}


In this part of the evaluation, we present the findings of our comparative analysis on the transferability of three adversarial attacks: FGSM, PGD and Momentum, conducted across four diverse datasets. These results offer insights into the effectiveness of these attacks, the vulnerability of different victim models and the choice of different surrogate models for each attack method.
the comparison is presented in Tables~\ref{tab:attacks_A} and~\ref{tab:attacks_B}, partitioned such that results for the CIFAR10 and ImageNet datasets appear in Table~\ref{tab:attacks_A}, and results for the X-ray and Road Sign datasets appear in Table~\ref{tab:attacks_B}.
The lower and upper bounds presented in the tables are from the PGD attack.

Overall we found that the HET ranking method is highly effective regardless of the attack algorithm used. In two of the datasets (ImageNet and CIFAR10) all three attack achieved near upper bound performance with HET. 

Interestingly, in our analysis we observed a high degree of transferability exhibited by the FGSM attack. Across these datasets, FGSM consistently demonstrated its effectiveness in crafting adversarial examples capable of successfully deceiving a range of diverse victim models.
The FGSM works by performing a single large perturbation on the image, unlike, the PGD and Momentum attacks which perform many small steps.
Using many attack steps may be preferable in a white box scenario since it allows targeting less prominent features in the victim model and thus creating less noticeable perturbations.
Nevertheless, in the black-box scenario, this might hinder the attack transferability, since these features might only be present in the surrogate model and not the victim.

Our experiments also shed light on the vulnerabilities of various victim models. Among them, \textit{EfficentNet}, a popular deep learning architecture, which was found to be the most susceptible victim model across all datasets. This discovery emphasizes the imperative need for robustness enhancements in \textit{EfficentNet} and similar models to mitigate the risks posed by adversarial attacks. Conversely, \textit{ViT},  and \textit{DenseNet121}, are proved to be reliant surrogate models, which provide better transferability capabilities compared to the others.
In the case of \textit{ViT}, this may be attributed to the transformer architecture that allows learning generic features, which do not necessarily reflect a certain architecture choice such as Convolutions.

Furthermore, our study highlighted the critical role played by the dataset characteristics in determining the success of adversarial attacks. Notably, ImageNet, with its large images and extensive class diversity comprising 1,000 classes, emerged as the most vulnerable dataset for adversarial attacks. The complexity and diversity inherent in ImageNet make it an enticing target for attackers, as it offers more opportunities to craft adversarial examples that can effectively deceive a wide range of victim models. These findings underscore the necessity for heightened security measures, particularly in complex and diverse datasets like ImageNet, to safeguard against adversarial threats.

\subsection{E2 - Perturbation Ranking}
\begin{table}[]
\resizebox{\linewidth}{!}{%
\begin{tabular}{|ccc|c|ccc|ccc|ccc|ccc|}
\hline
\multicolumn{3}{|c|}{}                                                                                                       & Lower B.                     & \multicolumn{3}{c|}{SoftMax}                                                               & \multicolumn{3}{c|}{SoftMax+Noise}                                                         & \multicolumn{3}{c|}{HET}                                                                   & \multicolumn{3}{c|}{Upper B.}                                                              \\ \cline{4-16} 
                                                  & Surrogate                                              & Victim          & ~                            & 5\%                          & 10\%                         & 20\%                         & 5\%                          & 10\%                         & 20\%                         & 5\%                          & 10\%                         & 20\%                         & 5\%                          & 10\%                         & 20\%                         \\ \hline
\multicolumn{1}{|c|}{}                            & \multicolumn{1}{c|}{}                                  & ViT\_l\_16      & \cellcolor[HTML]{FCBF7B}0.40 & \cellcolor[HTML]{FCBF7B}0.40 & \cellcolor[HTML]{FFEB84}0.60 & \cellcolor[HTML]{FDCA7D}0.45 & \cellcolor[HTML]{FA9473}0.20 & \cellcolor[HTML]{FBAA77}0.30 & \cellcolor[HTML]{FBAA77}0.30 & \cellcolor[HTML]{63BE7B}1.00 & \cellcolor[HTML]{63BE7B}1.00 & \cellcolor[HTML]{77C47D}0.95 & \cellcolor[HTML]{63BE7B}1.00 & \cellcolor[HTML]{63BE7B}1.00 & \cellcolor[HTML]{63BE7B}1.00 \\
\hline
\end{tabular}
}
\caption{The average transferability of a sample when ranking its potential perturbations for the CIFAR10 and ImageNet datasets over 100 trials. Columns represent the various ranking methods, and rows indicate the combination of victim and surrogate model architectures, ensuring that \( F_0 \) is chosen from architectures different from the victim's.}\label{tab:E2.1A}
\end{table}
\begin{table}[]
\resizebox{\linewidth}{!}{%

\begin{tabular}{|ccc|c|ccc|ccc|ccc|ccc|}
\hline
\multicolumn{3}{|c|}{}                                                                                                        &                              & \multicolumn{3}{c|}{SoftMax}                                                               & \multicolumn{3}{c|}{SoftMax}                                                               & \multicolumn{3}{c|}{HET}                                                                   & \multicolumn{3}{c|}{Upper B.}                                                              \\ \cline{5-16} 
                                                   & Surrogate                                              & Victim          & \multirow{-2}{*}{Lower B.}   & 5\%                          & 10\%                         & 20\%                         & 5\%                          & 10\%                         & 20\%                         & 5\%                          & 10\%                         & 20\%                         & 5\%                          & 10\%                         & 20\%                         \\ \hline
\multicolumn{1}{|c|}{}                             & \multicolumn{1}{c|}{}                                  & ViT             & \cellcolor[HTML]{C9DC81}0.61 & \cellcolor[HTML]{97CD7E}0.80 & \cellcolor[HTML]{B1D580}0.70 & \cellcolor[HTML]{BED981}0.65 & \cellcolor[HTML]{97CD7E}0.80 & \cellcolor[HTML]{B1D580}0.70 & \cellcolor[HTML]{B1D580}0.70 & \cellcolor[HTML]{63BE7B}1.00 & \cellcolor[HTML]{63BE7B}1.00 & \cellcolor[HTML]{63BE7B}1.00 & \cellcolor[HTML]{63BE7B}1.00 & \cellcolor[HTML]{63BE7B}1.00 & \cellcolor[HTML]{63BE7B}1.00 \\
\hline
\end{tabular}
}
\caption{The average transferability of a sample when ranking its potential perturbations for the X-Ray and Road Sign datasets over 100 trials. Columns represent the various ranking methods, and rows indicate the combination of victim and surrogate model architectures, ensuring that \( F_0 \) is chosen from architectures different from the victim's.}\label{tab:E2.1B}
\end{table}

\begin{figure*}
    \centering
    \begin{tabular}{c}
        ImageNet\\
        \includegraphics[width=\columnwidth]{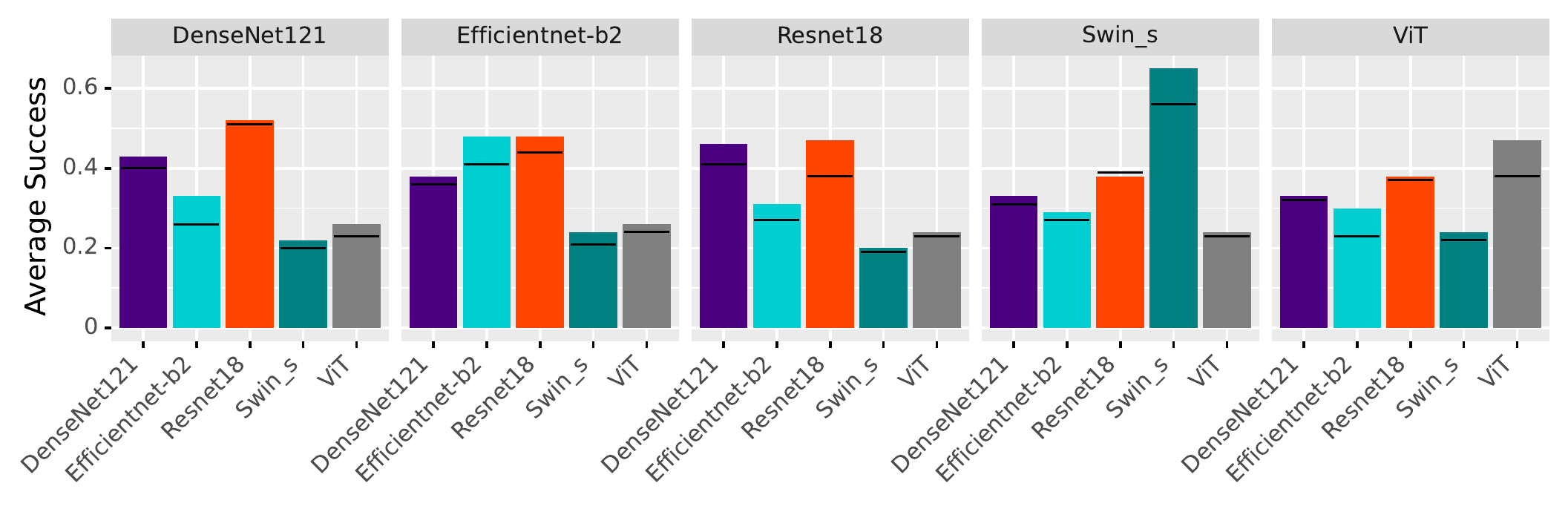}\vspace{-1.5em} \\
        CIFAR10\\
        \includegraphics[width=\columnwidth]{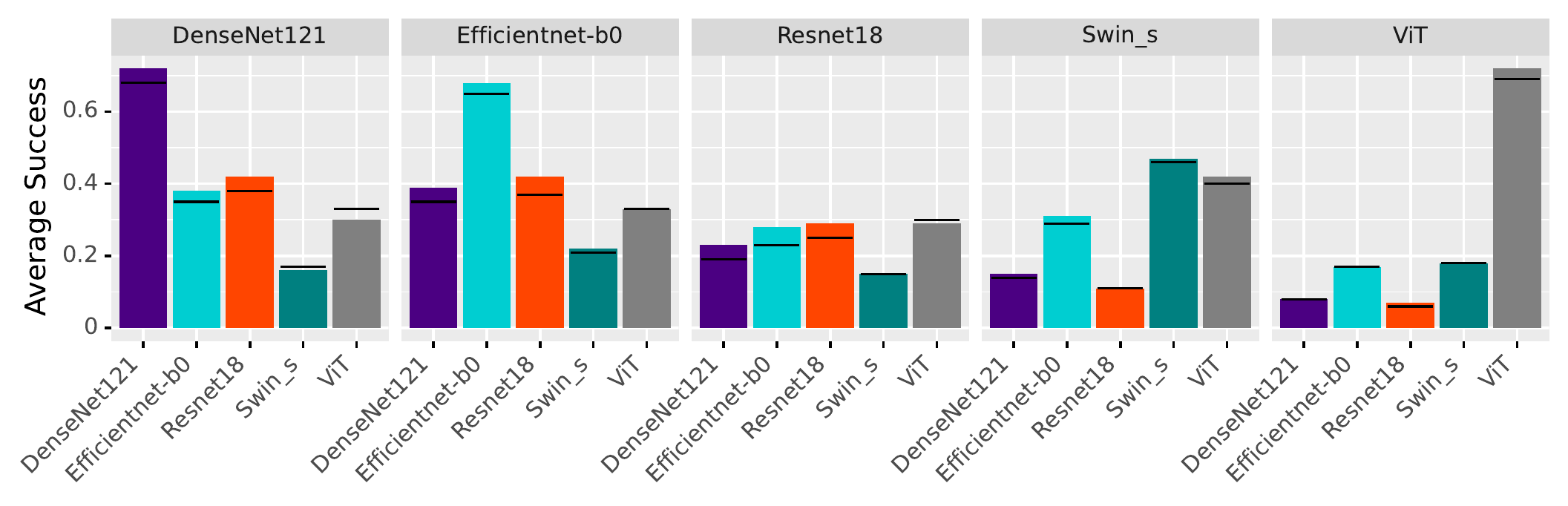}\vspace{-1.5em} \\
        Road Sign\\
        \includegraphics[width=\columnwidth]{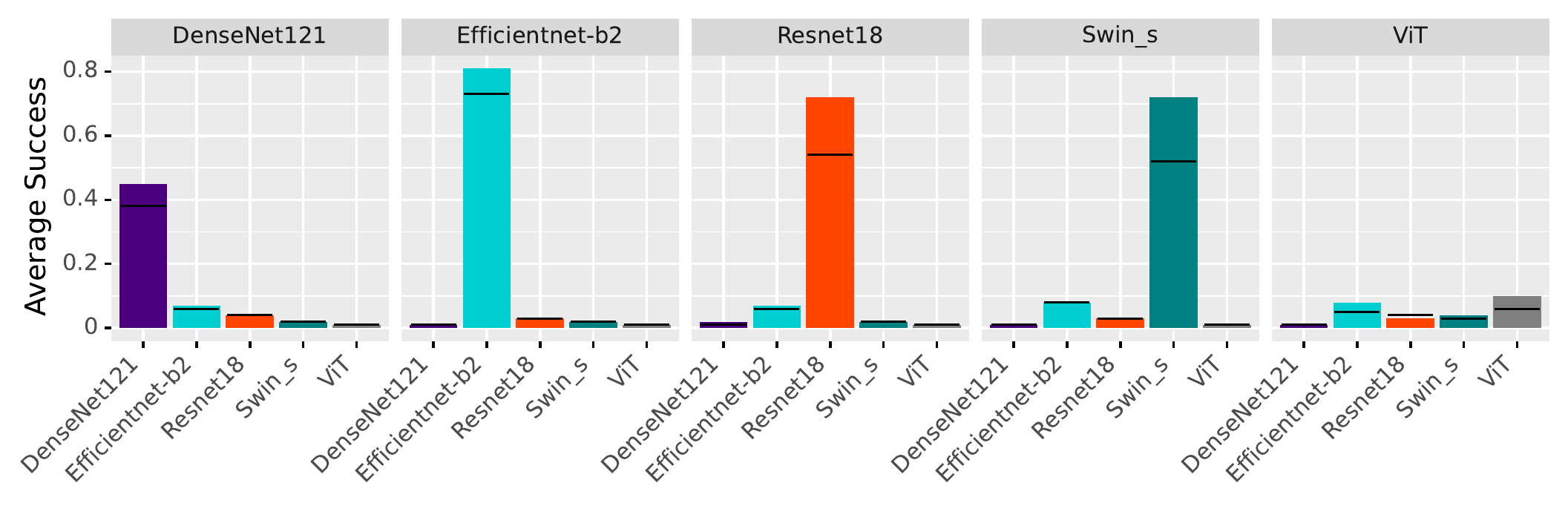}\vspace{-1.5em} \\
        X-Ray\\
        \includegraphics[width=\columnwidth]{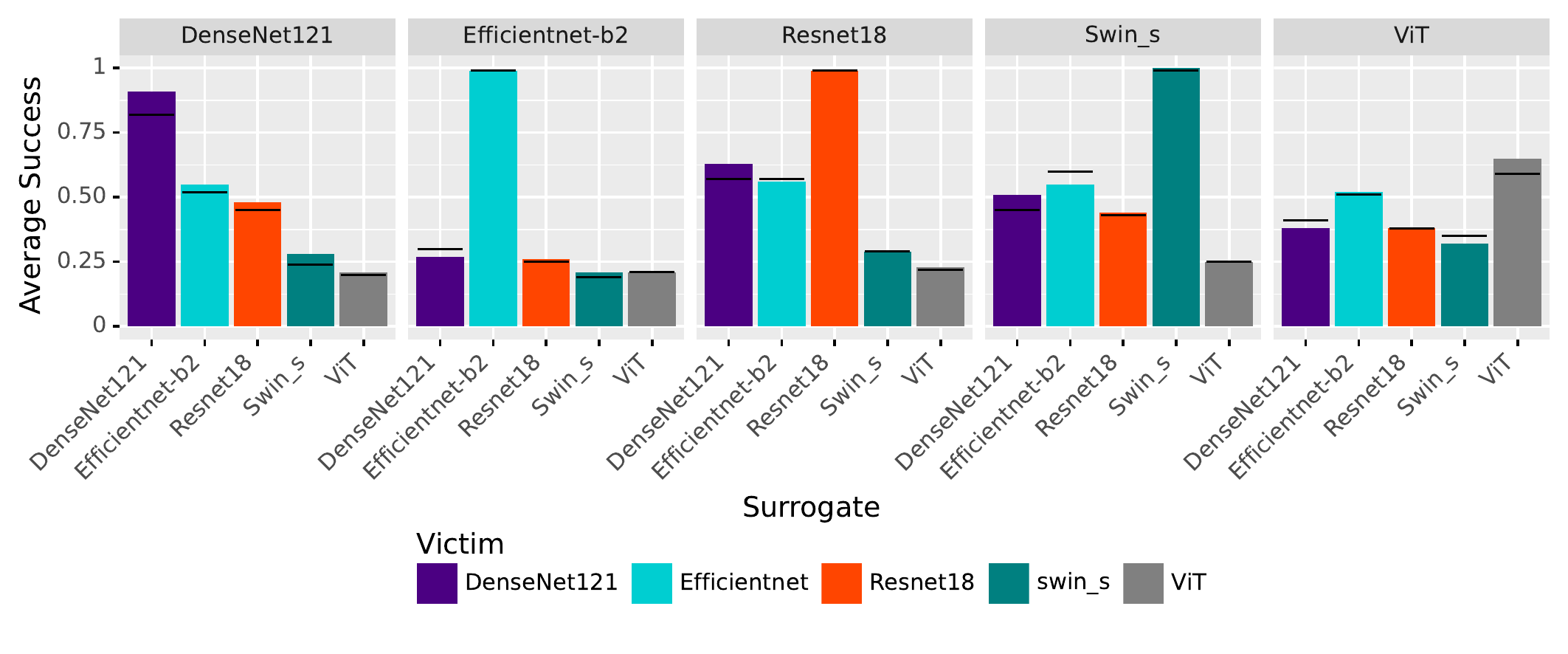}\vspace{-2em} \\
    \end{tabular}
    \caption{The average attack success rate when using the best pertubation ($k=1$). The rows are the datasets and the columns are the the architecture used for the surrogate $f'$. Bars indicate baseline results (no ranking).}
    \label{fig:pertubation_ranking}
\end{figure*}

\subsubsection{E2.1 - Performance.}

In Fig. \ref{fig:pertubation_ranking} we present the average transferability success rate of a sample when the highest ranked perturbation (out of 25) is used. Although perturbation ranking is less effective than image ranking, the figure shows that transferability can indeed be improved modestly in many situations. We also note that the attacker receives the largest benefit from ranking if the surrogate happens to be the same architecture as the victim. However, these cases can be considered rare in a strict blackbox setting.

We also observe that, similar to the results in E1.1, the road sign dataset is challenging to  perform ranking on. We believe this is because the road sign dataset is relatively small resulting in surrogate models with unaligned loss surfaces \cite{demontis2019adversarial}. However, when the attacker has a large enough dataset (disjoint from the victim) then ranking is effective (e.g., for the case of ImageNet, CIFAR10 and X-Ray).

In Tables \ref{tab:E2.1A} and \ref{tab:E2.1B} we present the results when \textit{ranking images} for different $k$ after applying the best perturbation to each image. Table \ref{tab:E2.1A} presents the findings for the CIFAR10 and ImageNet datasets, while Table \ref{tab:E2.1B} provides the results for the X-Ray and Road Sign datasets. Each cell within these tables indicates the average transferability from 100 random images selected from the dataset, with the columns denoting the ranking methods alongside the established upper and lower bounds. The rows detail the combinations of architectures for the victim and surrogate models.

The results underscore the proficiency of HET in consistently pinpointing the most transferable perturbation for a given sample. The significance of this capability is highlighted by the comparison to the lower bound of transferability—averaging at or below 30\% across the datasets—which HET substantially elevates to an average of 70\% or greater. Particularly striking is HET's performance for the ImageNet, X-Ray, and Road Sign datasets, where it nearly mirrors the upper bound. In the X-Ray and ImageNet datasets, HET achieves perfect transferability across almost all black-box scenarios (selection of architecture combinations).

Even as the value of \(k\) increases, representing a broader selection of top-ranking perturbations, HET maintains its superior performance. It exhibits an enhancement of up to 60\% in transferability over the lower bound for larger values of \(k\). This improvement is noteworthy, demonstrating the robustness of HET in a variety of conditions.

Conversely, the baseline methods of SoftMax and SoftMax+Noise generally hover around the lower bound, occasionally achieving up to a 40\% increase in transferability, yet still falling short of the performance attained by HET. The disparity between these methods and HET is particularly evident within the Road Sign dataset. For lower values of \(k\), the baselines are analogous to the lower bound, whereas HET's results are akin to the upper bound, accentuating the substantial advantage provided by the HET ranking strategy in these scenarios.

\section{Related Work}
There has been a great amount of research done on adversarial transferability, discussing attacks \cite{naseer2019cross,wang2021admix,Springer2021ALR,zhu2021rethinking}, defences \cite{guo2018countering,madry2018towards} and performing general analysis of the phenomena \cite{tramer2017space,katzir2021s,demontis2019adversarial}. However these works do not rank transferability from the attacker's side, but rather evaluate transferability of entire datasets directly on the victim model. In contrast, our work defines the task of transferability ranking for blackbox attackers and proposes methods for performing the task.  

Some works have proposed techniques and measures for identifying samples that transfer better than others. For example, \cite{sourcesamples} show that some subsets of samples transfer better than others and suggest ways to curate evaluation datasets for transferability. They found that there is a connection between the transferability success of $x'$ to $f$ when observing the Softmax outputs of $f(x)$.
In an other work, sensitivity of $x$ to Gaussian noise on $f$ was found to be correlated with transferability of $x'$ to $f$ \cite{zhu2021rethinking}. 
However, in contrast to our work, the processes described in these works only apply to whitebox scenarios with access to $f$. In many cases, attackers do not have access to the victim's model, cannot send many adversarial examples to the model without raising suspicion, or cannot receive feedback from the model (e.g., classifiers used in airport x-ray machines \cite{akcay2022towards}). Therefore, they cannot evaluate their attack on the victim's model prior. Moreover these works do not define the task of transferability ranking or suggest methods for measuring transferability such as \textit{transferability at $k$}.

Finally, in contrast to prior works, we suggest a more grounded approach to evaluating model security in transfer attacks. We recommend that the community evaluate their models against the top $k$ most transferable samples from a blackbox perspective, and not by taking the average success of all samples in whitebox perspective. This is because a true attacker will likely select the best samples to use with ultimately increases the performance (threat) of transferability attacks.

\section{Conclusion}
The results garnered from our extensive experimentation with the HET ranking strategy offer compelling evidence of its effectiveness in enhancing the transferability of adversarial examples. Notably, the strategy demonstrates remarkable efficacy in the context of improving the transferability of a single specific sample, a scenario of particular relevance to black-box attackers. The findings reveal that attackers are not `stuck' with a resulting adversarial example and its likelihood of transferability; rather, by applying HET to select the best perturbation from a set of trials, attackers can significantly boost the likelihood of a successful transfer. Ideal results from transferability ranking can only be achieved when the attacker happens to guess the victim's architecture and use it to generate the adversarial examples. However, our results show that if the attacker only needs to send less than 10\% of a set of potential samples, then ranking increases the chance of transferability for those samples significantly, and nearly guarantees it when selecting only one ($k=1$).

This efficiency in attack methodology is a critical advantage in practical adversarial settings. It underscores a reduced risk for the attacker, as there is no longer a requirement to send to the victim a vast number of perturbations in the hope of stumbling upon a successful one. Instead, HET provides a systematic and predictive approach to identifying vulnerabilities in the victim \textbf{without sending any samples to the victim first}.

Moreover, the consistent performance of HET across various datasets and model architectures offers a deeper understanding of the intrinsic characteristics of adversarial attacks. As suggested in previous works, it supports the claim that that different models, despite their distinct architectural designs, often share common weaknesses~\cite{naseer2019cross,wang2021admix,Springer2021ALR,zhu2021rethinking}.

The ability to predict and exploit these shared vulnerabilities is significant. It implies that there is an underlying structure to the adversarial space that HET can effectively navigate. By selecting perturbations that are universally potent across models, HET enables attackers to reliably anticipate the success of their attacks in black-box settings. This observation not only affirms the utility of HET but also provides a valuable insight into the nature of transferability, suggesting that the successful perturbation is not a fluke but a systematic exploitation of a model's fundamental susceptibilities. The implications of this for both attackers and defenders are profound, as it calls for a deeper exploration into the robustness of models and the development of more sophisticated defense mechanisms.

\begin{acks}
This material is based upon work supported by the Zuckerman STEM Leadership Program. This project has received funding from the European union's Horizon 2020 research and innovation programme under grant agreement 952172.

\begin{figure}[h]
    \centering
    \includegraphics[width=0.1\textwidth]{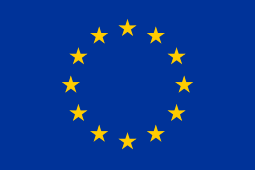}
\end{figure}
\end{acks}

\bibliographystyle{ACM-Reference-Format}
\bibliography{egbib}

\end{document}